\documentclass{article}

\PassOptionsToPackage{numbers,compress}{natbib}
\PassOptionsToPackage{numbers,compress}{natbib}
\usepackage[preprint]{neurips_2026}

\usepackage[utf8]{inputenc} 
\usepackage[T1]{fontenc}    
\usepackage{hyperref}       
\usepackage{url}            
\usepackage{booktabs}       
\usepackage{amsfonts}       
\usepackage{nicefrac}       
\usepackage{microtype}  
\usepackage{natbib}   
\usepackage{xcolor}         
\usepackage{amsmath}
\usepackage{amssymb}

\usepackage{amsthm}
\usepackage{float}

\theoremstyle{plain}
\newtheorem{theorem}{Theorem}[section]
\newtheorem{lemma}[theorem]{Lemma}

\theoremstyle{remark}

\usepackage{algorithm}
\usepackage{algpseudocode}
\usepackage{graphicx}
\newcommand{\eatme}[1]{ }
\title{CDS: Counterfactual Directionality Score for Structured Interventions in Spatial Graphs}

\author{%
  Humaira Anzum \\
  Md Ishtyaq Mahmud \\
  Jagan Mohan Reddy Dwarampudi \\
  Tania Banerjee\\
  Department of Electrical and Computer Engineering \\
  University of Houston, Houston, TX, USA \\
  \texttt{humaira.snigdha.22@gmail.com, ishtyaqtushar@gmail.com} \\
  \texttt{jdwaramp@cougarnet.uh.edu, tbanerjee@uh.edu} \\
}

\begin{document}

\maketitle

\begin{abstract}
Quantifying directional influence between node populations is a fundamental problem in graph-based modeling, particularly in spatial biological systems where cell–cell interactions shape functional outcomes. Existing approaches based on attention, attribution, or correlation capture associations but do not provide a principled framework for evaluating directional effects under controlled perturbations. We introduce a framework for structured counterfactual interventions in graph-based models to estimate directional influence between node types. Our approach trains a Neighbor Influence Model (NIM) to predict node states from local neighborhoods and applies constrained interventions that modify neighborhood composition while preserving key spatial and structural properties. We define the Counterfactual Directionality Score (CDS), which measures the change in predicted node state induced by targeted perturbations, and provide a theoretical interpretation of CDS as a finite-difference
measure of local intervention sensitivity. To obtain valid uncertainty estimates, we introduce a core-level bootstrap procedure that accounts for dependencies within spatial samples. Experiments on synthetic spatial graphs with known directional structure show that CDS recovers directional influence, remains well calibrated under null conditions, and is robust to confounding signals, while preliminary results on spatial transcriptomics data reveal biologically plausible and consistent interactions across tissue cores.
\end{abstract}

\section{Introduction}

Spatially-resolved single-cell technologies reveal how cells interact within tissues. A central question is \emph{directional influence}: how does one cell type affect the molecular state of another? Answering this question is key to understanding tumor–immune interactions, stromal remodeling, and signaling in health and disease. Graph neural networks (GNNs) model such data by representing cells as nodes and spatial proximity as edges. While GNNs achieve strong predictive performance, they provide limited tools for quantifying directional effects between cell populations. Existing approaches, such as attention weights, feature attribution, and correlation-based cell–cell interaction analyses, capture associations but often fail to distinguish true directional influence from confounding factors such as shared microenvironmental signals or global expression trends.

We introduce a framework for \emph{structured counterfactual interventions} in graph-based models to estimate directional influence. Our principle is: if a sender population influences a receiver’s state, systematically removing that population should change the receiver’s predicted cellular state. We operationalize this idea by training a \emph{Neighbor Influence Model (NIM)} that predicts a cell’s transcriptomic state from its spatial neighborhood. Then we apply counterfactual interventions that modify neighborhood composition while preserving key spatial and structural constraints, including neighborhood size, spatial weighting, and tissue context. The resulting \emph{Counterfactual Directionality Score (CDS)} quantifies the sensitivity of a receiver’s predicted state to targeted perturbations. We further introduce within‑type interventions as an empirical baseline to distinguish the effects of cell‑type presence from individual cell‑state variability. We provide a theoretical interpretation of CDS as a finite-difference measure of local intervention sensitivity under structured neighborhood perturbations. Our structured interventions isolate compositional effects while controlling for structural confounders. To obtain valid uncertainty estimates, we develop a core‑level bootstrap procedure that accounts for dependencies within tissue samples.

On synthetic spatial graphs with known directional structure, CDS recovers ground truth directional structure, remains well calibrated under null conditions, and is robust to confounding signals. Compared to baseline methods, CDS uniquely combines sensitivity to true interactions with robustness to spurious correlations. Preliminary results on spatial transcriptomics datasets show that CDS identifies biologically plausible sender–receiver relationships with consistent behavior across tissue cores.

\paragraph{Contributions.}
Our main contributions are: (1) a framework for structured
counterfactual interventions in spatial graphs that preserves key
structural properties; (2) CDS, a model-based measure of directional
influence between cell types; (3) a theoretical characterization of CDS
as a finite-difference intervention measure under structured
perturbations; (4) a core-level bootstrap procedure for statistically
principled inference; and (5) empirical validation showing that CDS
recovers directional structure, remains well calibrated under null
conditions, and is robust to confounding.

\section{Related Work}
Our work connects several research areas: interpretability for graph neural
networks, attention-based models, computational methods for cell–cell
interaction inference, and causal machine learning with counterfactual
reasoning.

Several methods have been developed to explain predictions of GNNs.
GNNExplainer~\cite{ying2019gnnexplainer} identifies compact subgraphs and node
features that are most influential for a GNN’s predictions.
PGExplainer~\cite{luo2020pgexplainer} provides a parameterized approach that
efficiently generates explanations for multiple instances. While these methods
reveal which graph components are important, they focus on attribution rather
than estimating directional effects between node populations. Our CDS explicitly perturbs neighbourhood
composition in a structured way to estimate directional influence.

Attention mechanisms in GNNs, such as Graph Attention Networks
(GATs)~\cite{velickovic2018graph}, learn adaptive weights over neighbours and
have been used to interpret which interactions are “important”. However,
attention weights reflect correlations within the observed data distribution
and do not provide a counterfactual characterization of directionality. Our
work instead uses explicit counterfactual interventions to isolate the effect
of a specific cell type’s presence.

In computational biology, tools such as CellPhoneDB~\cite{efremova2020cellphonedb}
and NicheNet~\cite{browaeys2020nichenet} infer ligand–receptor interactions
from single-cell and spatial transcriptomics data using prior knowledge of
molecular interactions. More recent spatial CCI
methods~\cite{armingol2021deciphering, liao2022spatalk} incorporate spatial
proximity to score interactions. Though these methods are powerful for
hypothesis generation, they are primarily correlation-based or rely on curated
databases. Our framework is model-agnostic and learns a predictive model of
cell state from neighbourhood composition. This enables principled
counterfactual estimation without requiring predefined interaction databases.

Causal inference provides a formal language for estimating intervention
effects~\cite{pearl2009causality}. In machine learning, counterfactual
explanations have been applied to image and tabular data
(e.g.,~\cite{wachter2017counterfactual}). For graphs, recent work has proposed
counterfactual graph generation for model explanation
(e.g.,~\cite{abid2022counterfactual}), but these approaches typically modify
graph structure globally or focus on graph-level classification. To our
knowledge, our work is the first to combine a learned graph-based surrogate
model with spatially constrained, cell-type-specific counterfactual
interventions to estimate directional influence between node populations in
spatial biological graphs. The core-level bootstrap further provides
statistically principled inference that respects the dependence structure of
tissue samples.

\section{Problem Setup\label{sec:problem}}
We consider a spatially-resolved single-cell dataset represented as a graph 
$G = (V, E)$, where each node $i \in V$ corresponds to a cell, and edges 
$(i,j) \in E$ encode spatial proximity between cells. Each cell is associated 
with (1) a feature vector $x_i \in \mathbb{R}^p$ representing its transcriptomic 
state, and (2) a discrete cell-type label $\tau(i) \in \mathcal{T}$. 
Let $X \in \mathbb{R}^{|V| \times p}$ denote the matrix of all cell features.

\paragraph{Spatial neighborhood.}
For each cell $i$, we define a local neighborhood $N(i) \subset V$ based on 
spatial proximity. In practice, $N(i)$ may be constructed via $k$-nearest 
neighbors in physical space, but our formulation is agnostic to the specific 
construction as long as it captures local cell--cell interactions.

\paragraph{Sender and receiver populations.}
For a pair of cell types $(s, r) \in \mathcal{T}^2$, we define the sender and 
receiver populations:
\begin{align}
    S &= \{ i \in V : \tau(i) = s \}, \\
    R &= \{ i \in V : \tau(i) = r \}.
\end{align}

\paragraph{Interaction-driven cell state.}
We view the observed transcriptomic state of a receiver cell $i \in R$ as the 
result of both intrinsic factors and extrinsic influences from its local 
neighborhood:
\begin{equation}
    x_i = f\big(x_i^{\mathrm{int}}, \{x_j : j \in N(i)\}\big) + \epsilon_i
\end{equation}
where $x_i^{\mathrm{int}}$ denotes intrinsic (cell-autonomous) components and 
the neighborhood term captures microenvironmental influence. $\epsilon_i$ captures unmodeled variability.

\paragraph{Problem objective.}
Our goal is to quantify the \emph{directional influence} of sender-type cells 
on receiver-type cells. Formally, for each receiver cell $i \in R$, we define a neighborhood composition variable $Z_i^{(s)}$ that summarizes the
contribution of sender-type cells in $N(i)$. We are interested in estimating the effect:
\begin{equation}
    \Delta_i^{(s \to r)} = 
    f\big(x_i^{\mathrm{int}}, Z_i^{(s)}\big) - 
    f\big(x_i^{\mathrm{int}}, 0\big),
\end{equation}
which represents the change in the receiver cell state induced by removing sender-type 
cells from the local neighborhood.

\paragraph{Scientific question.}
This formulation allows us to address the central question:
\emph{Do neighboring sender-type cells influence the transcriptomic
program of receiver-type cells, and if so, what is the magnitude and
direction of this effect?}


\section{Neighbor Influence Model}

\subsection{Distance-Weighted Neighborhood Aggregation}

We model the influence of a cell's local microenvironment through a
distance-weighted aggregation of its neighbors. For each cell $i$ and
neighbor $j \in N(i)$, let $d_{ij}$ denote their Euclidean distance
in tissue space. We define normalized weights:
\begin{equation}
w_{ij}
= \frac{\exp\bigl(-d_{ij} / \alpha\bigr)}
{\sum_{j' \in N(i)} \exp\bigl(-d_{ij'} / \alpha\bigr)},
\label{eq:softmax_weights_clean}
\end{equation}
where $\alpha > 0$ controls the spatial decay of influence.

Using these weights, we define the neighborhood representation
$a_i \in \mathbb{R}^p$:
\begin{equation}
a_i = \sum_{j \in N(i)} w_{ij} x_j.
\end{equation}

\subsection{Modeling Neighborhood Influence}

We approximate the unknown function $f$ introduced in
Section~\ref{sec:problem} using a graph-based parametric model
$f_\theta$, referred to as NIM:
\begin{equation}
\hat{z}_i = f_\theta(a_i, \tau(i)) \in \mathbb{R}^p.
\label{eq:nim_output_clean}
\end{equation}

The model takes as input the neighborhood representation $a_i$ and the
receiver cell type $\tau(i)$, and predicts the receiver’s transcriptomic
state. Learning $f_\theta$ provides a data-driven estimate of how local
neighborhood composition maps to cellular gene expression.

\paragraph{Architecture.}
$f_\theta$ is implemented as a multi-layer feedforward network with
residual connections. The input $a_i$ is projected into a latent space,
processed by $L$ residual blocks, and modulated by a
cell-type-conditioned gating mechanism. Specifically, each cell type
$\tau(i)$ is embedded into a vector $e_{\tau(i)}$, which is transformed into a
feature-wise gate:
\begin{equation}
g_i = \sigma(W_{\mathrm{type}} e_{\tau(i)} + b_{\mathrm{type}}),
\end{equation}
that multiplicatively modulates the hidden representation.

This gating allows the model to capture heterogeneous response
functions, i.e., different receiver cell types may respond differently
to the same neighborhood signal.

\subsection{Training Objective}

We train $f_\theta$ to minimize the discrepancy between predictions
$\hat{z}_i$ and observed states $x_i$ using the Huber loss:
\begin{equation}
\mathcal{L}(\theta)
= \frac{1}{|\mathcal{B}|} \sum_{i \in \mathcal{B}}
\ell_\delta(\hat{z}_i, x_i),
\end{equation}
where $\ell_\delta$ is defined as the Huber loss.

\paragraph{Interpretation.}
A low prediction error implies that neighborhood composition contains
predictive information about the receiver cell state. Importantly, this
learned mapping $f_\theta$ serves as a surrogate model for evaluating
counterfactual neighborhood perturbations (Section~\ref{sec:theory}), enabling estimation of sensitivity to sender-to-receiver perturbations.

\section{CDS}

\subsection{Intuition}

While NIM captures how neighborhood
composition predicts cell state, it does not by itself identify which
neighboring cell types drive this effect. To address this, we introduce \textbf{CDS}, which measures
the sensitivity of a receiver cell's predicted state to targeted
perturbations of its neighborhood.

\subsection{Counterfactual Predictions}

Let $f_\theta$ be the trained NIM. For a receiver cell $i$, the original
prediction is:
\begin{equation}
\hat{z}_i = f_\theta(a_i, \tau(i)),
\end{equation}
where $a_i$ is the original neighborhood aggregation.

We construct a counterfactual neighborhood $N^{\mathrm{cf}}(i)$ via an
intervention operator $\mathcal{I}_{S \rightarrow R}$ that modifies the
presence of sender-type cells while preserving the spatial weighting
structure. The corresponding counterfactual prediction is:
\begin{equation}
\hat{z}_i^{\mathrm{cf}} = f_\theta(a_i^{\mathrm{cf}}, \tau(i)),
\end{equation}
where $a_i^{\mathrm{cf}}$ denotes the neighborhood aggregation computed
from the counterfactual neighborhood $N^{\mathrm{cf}}(i)$.

\subsection{Intervention Operators}

We define two structured intervention operators that modify neighborhood
composition while preserving key spatial and structural properties.
These interventions are designed to isolate the contribution of
sender-type cells to receiver predictions.

\paragraph{Type-Swap Intervention.}
This operator removes the presence of sender-type cells from the
receiver’s neighborhood. For each receiver $i \in R$ and each neighbor
$j \in N(i)$ with $\tau(j) = s$, we replace $j$ with a cell
$j'$ sampled from the same tissue core such that $\tau(j') \neq s$.

To ensure physically plausible perturbations, we perform
\emph{distance-bin-preserving replacement}. The $k$ neighbor slots are
partitioned into spatial bins based on distance, and each sender slot is
preferentially replaced by a non-sender cell from the same bin; if no
such candidate exists, the search expands to nearby bins.

This construction preserves three structural invariants:
\begin{enumerate}
    \item Degree preservation: $|N^{\mathrm{cf}}(i)| = |N(i)| = k$.
    \item Microenvironment consistency: replacement cells are
    drawn from the same tissue core as $i$.
    \item Distance-slot preservation: spatial weights
    $\{w_{ij}\}$ remain unchanged, so differences between $a_i$ and
    $a_i^{\mathrm{cf}}$ arise solely from changes in neighborhood composition.
\end{enumerate}

As a result, changes in prediction are attributable primarily to changes in neighborhood composition rather than artifacts of graph structure or spatial weighting.

\paragraph{Within-Type Intervention.}
This operator help characterize the effect of \emph{sender cell identity} while
preserving sender cell type. For each sender neighbor $j \in N(i)$ with
$\tau(j) = s$, we replace $j$ with a different cell $j'$ of the same
type, drawn from the donor pool:
\begin{equation}
    \mathcal{D}_i
    = \bigl\{\, j' \in V :
        \tau(j') = s,\;
        \mathrm{core}(j') = \mathrm{core}(i),\;
        j' \neq i
    \,\bigr\}.
\end{equation}

If $\mathcal{D}_i = \emptyset$, the receiver $i$ is excluded from
analysis. This intervention preserves the presence of sender-type cells
while randomizing which specific cells occupy neighborhood slots.

\subsection{CDS Definition}

For each receiver cell $i$, we define:
\begin{equation}
\mathrm{CDS}_i
= \frac{1}{p} \left\| \hat{z}_i^{\mathrm{cf}} - \hat{z}_i \right\|_1.
\end{equation}
This measures the average change in predicted state under the
counterfactual intervention.

A larger $\mathrm{CDS}_i$ indicates that the perturbed cell type
substantially influences the receiver's transcriptomic state.

We also define a signed variant:
\begin{equation}
\mathrm{CDS}_i^{\mathrm{sgn}}
= \frac{1}{p} \sum_{m=1}^{p}
\bigl(\hat{z}_{i,m}^{\mathrm{cf}} - \hat{z}_{i,m}\bigr),
\end{equation}
which captures the direction of influence (upregulation vs.\ suppression).

Finally, we aggregate over receiver cells:
\begin{equation}
\mathrm{CDS}_{S \rightarrow R}
= \frac{1}{|R^*|} \sum_{i \in R^*} \mathrm{CDS}_i,
\end{equation}
where $R^*$ includes receivers for which the intervention is valid.

\paragraph{Interpretation.}
CDS provides a model-based measure of directional influence: it
quantifies how much the predicted state of receiver cells depends on
the presence or identity of sender-type cells in their local
neighborhood.

\section{Theoretical Properties of Counterfactual Directionality\label{sec:theory}}

CDS can be interpreted as a finite-difference measure of the sensitivity
of model predictions to structured neighborhood perturbations. The
difference $a_i^{\mathrm{cf}} - a_i$ captures changes in neighborhood
composition induced by the intervention while preserving spatial
structure and neighborhood size.

Structured interventions preserve neighborhood size, spatial distances
(and therefore weights $w_{ij}$), and receiver type, so
\[
a_i^{\mathrm{cf}} - a_i
=
\sum_{m \in N(i)}
w_{im}
\bigl(x_m^{\mathrm{cf}} - x_m\bigr),
\]
thereby separating compositional changes from changes in graph topology.

Under within-type exchangeability,
\[
\mathbb{E}[a_i^{\mathrm{cf}} - a_i] = 0,
\]
providing an empirical null baseline. Because CDS depends on the aggregated neighborhood representation $a_i$,
it captures only effects mediated through this representation. If $f_\theta$ is $L$-Lipschitz, then
\[
\mathrm{CDS}_i
\leq
\frac{L}{p}
\|a_i^{\mathrm{cf}} - a_i\|_2.
\]

For statistical inference, we use a core-level bootstrap procedure:
we resample $M$ tissue cores with replacement, compute bootstrap means
$\mu^{(b)}$, and construct percentile confidence intervals to account
for dependence within tissue cores.

\paragraph{Implementation.}
We train NIM using standard optimization procedures and compute CDS via
structured counterfactual perturbations of local neighborhoods. Full
algorithmic details are provided in Appendix~\ref{sec:a}.

\section{Experiments\label{sec:expt}}

We evaluate whether CDS recovers directional influence under controlled synthetic settings and compare it against standard attribution and perturbation baselines. Our evaluation focuses on directionality, robustness to confounding, and stability. These experiments empirically validate the theoretical properties of CDS derived in Section~\ref{sec:theory}, including its sensitivity to neighborhood perturbations and robustness under null exchangeability.

\subsection{Experimental Setup}

\paragraph{Synthetic Spatial Generator}
We construct spatial graphs with three cell types: sender ($S$), receiver ($R$), and background ($B$). Cell coordinates are sampled uniformly and a $k$-nearest neighbor graph is built with distance-based softmax weights.

Receiver features are generated under three regimes:

\begin{itemize}
\item \textbf{Positive (true influence):}
\[
z_i = \mu_R + a_i W + \epsilon,
\]
where $\mu_R \in \mathbb{R}^p$ is a receiver-type baseline expression
vector, $a_i$ is the weighted aggregation of neighboring sender
features, $W \in \mathbb{R}^{p \times p}$ controls sender-to-receiver
influence strength, and $\epsilon \sim \mathcal{N}(0,\sigma^2 I)$ is
Gaussian noise.

\item \textbf{Null (no influence):}
\[
z_i \sim \mathcal{N}(\mu_R, \sigma^2 I),
\]
independent of neighboring cells.

\item \textbf{Spurious (confounded):}
A shared latent variable induces correlation across cell types without local directional influence.
\end{itemize}

This design preserves feature marginals while controlling the underlying directional dependency structure.

\paragraph{Predictive Model}
We train NIM, a residual neural network that predicts cell features from neighborhood aggregates with cell-type gating. CDS is computed by comparing model predictions under original and counterfactual neighborhoods constructed via constrained neighbor replacement.

\subsection{Synthetic Validation}

\begin{figure}[t]
    \centering
    \includegraphics[width=\linewidth]{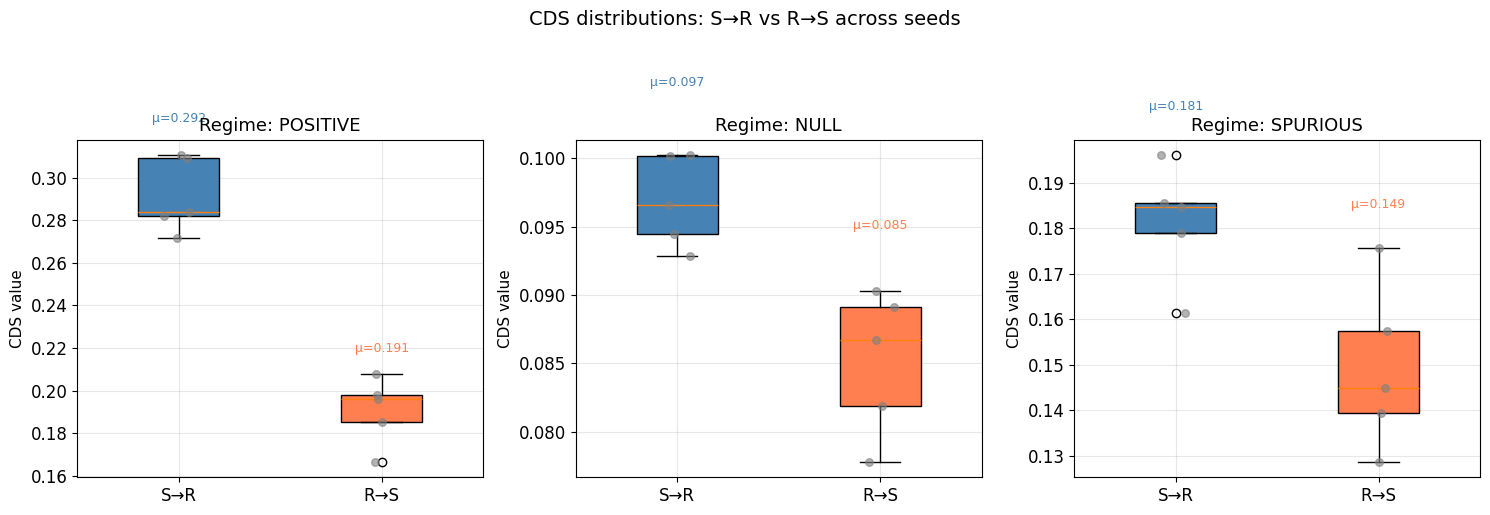}
    \caption{\textbf{Directional CDS across synthetic regimes.}
    Boxplots comparing $\mathrm{CDS}_{S \rightarrow R}$ and $\mathrm{CDS}_{R \rightarrow S}$ in positive, null, and spurious regimes.}
    \label{fig:cds-directionality}
\end{figure}

\paragraph{Directional Recovery}
We first evaluate whether CDS correctly recovers directional structure under controlled synthetic settings. Figure~\ref{fig:cds-directionality} shows that in the positive regime, CDS consistently assigns higher values to the true direction ($S \rightarrow R$) than the reverse ($R \rightarrow S$) across all noise levels and random seeds. This gap remains stable as noise increases, indicating that the learned model captures directional dependencies rather than symmetric correlations. In contrast, in the null regime, both directions yield near-zero CDS values, confirming that the method does not produce spurious directional signal (Table~\ref{tab:main-results}) and is well calibrated in the absence of true interaction.

\paragraph{Null Control}

Under the null setting, where receiver features are independent of neighboring sender cells, CDS values collapse to near zero for both directions (Table~\ref{tab:main-results}), confirming that the method does not introduce spurious directional signal under null conditions.

\paragraph{Spurious Correlation Control}

To test robustness to confounding, we introduce a shared latent signal affecting all cell types. As shown in Figure~\ref{fig:cds-directionality} and Table~\ref{tab:main-results}, CDS values remain lower than in the positive regime, and the directional gap is reduced.

This indicates that CDS preferentially captures structured local influence rather than global correlations, and does not incorrectly attribute strong directional effects under confounding.

\subsection{Quantitative Evaluation}

\begin{figure}[h]
    \centering
    \includegraphics[width=0.7\linewidth]{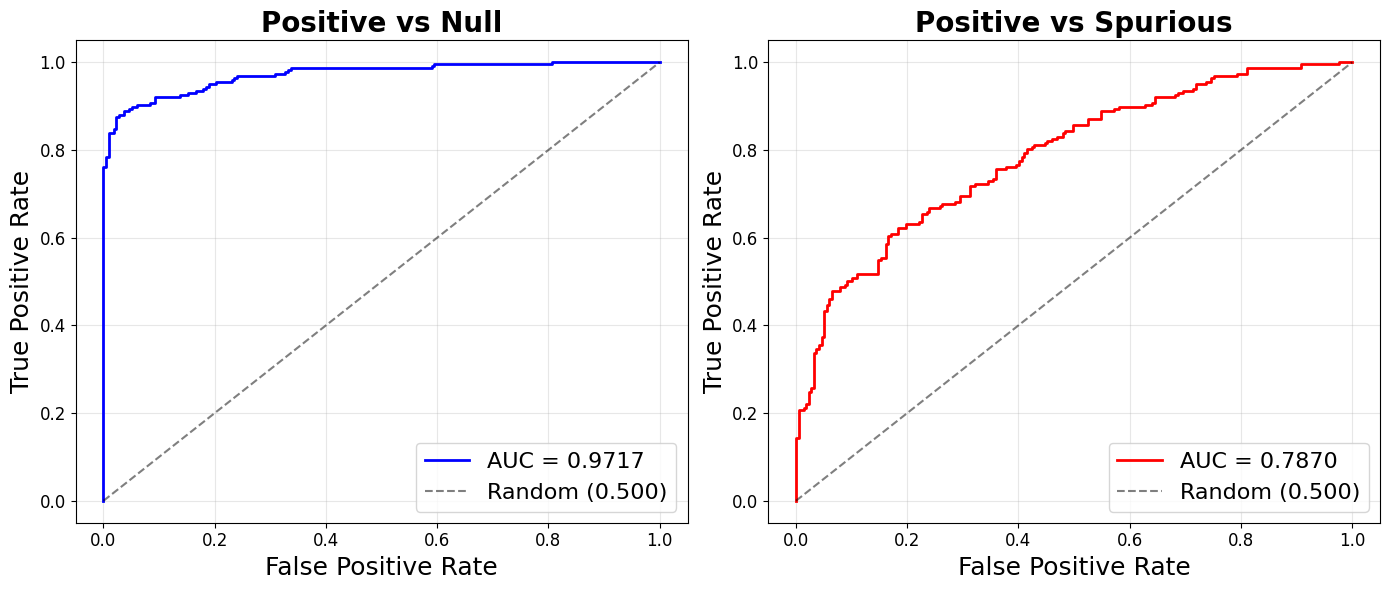}
    \caption{\textbf{Influence detection performance.}
    ROC curves for distinguishing positive influence from null and spurious controls using CDS scores.}
    \label{fig:roc-curves}
\end{figure}

We evaluate the ability of CDS to distinguish true interaction from null and confounded settings using ROC analysis. As shown in Figure~\ref{fig:roc-curves}, CDS achieves high AUC for distinguishing positive from null regimes (AUC = 0.97) and maintains above-chance discrimination under confounding (AUC = 0.79). These results indicate that CDS captures structured directional signals beyond simple correlation-based separation. Table~\ref{tab:main-results} summarizes CDS magnitude and stability across regimes. In the positive setting, CDS is substantially larger than in null and spurious regimes, with a low coefficient of variation across bootstrap samples. Bootstrap analysis further shows that CDS estimates are tightly concentrated, with narrow confidence intervals, indicating stable estimation across tissue cores. We compare CDS against random perturbation and gradient sensitivity baselines (Table~\ref{tab:baseline-comparison}). CDS consistently achieves higher scores in the positive regime while remaining near zero under null conditions.

\begin{table}[h]
\centering
\footnotesize
\begin{tabular}{lcccc}
\toprule
\textbf{Regime} & \textbf{CDS (S→R)} & \textbf{CDS (R→S)} & \textbf{AUC vs. Positive} & \textbf{Stability (CV)} \\
\midrule
Positive  & $0.2915 \pm 0.0156$ & $0.1907 \pm 0.0140$ & ---   & $0.034 \pm 0.016$ \\
Null      & $0.0969 \pm 0.0030$ & $0.0852 \pm 0.0047$ & 0.970 & $0.023 \pm 0.009$ \\
Spurious  & $0.1813 \pm 0.0114$ & $0.1492 \pm 0.0161$ & 0.793 & $0.055 \pm 0.014$ \\
\bottomrule
\end{tabular}
\caption{\textbf{Summary of CDS performance across regimes.}}

\label{tab:main-results}
\end{table}

\eatme{
\begin{figure}[h]
    \centering
    \includegraphics[width=\linewidth]{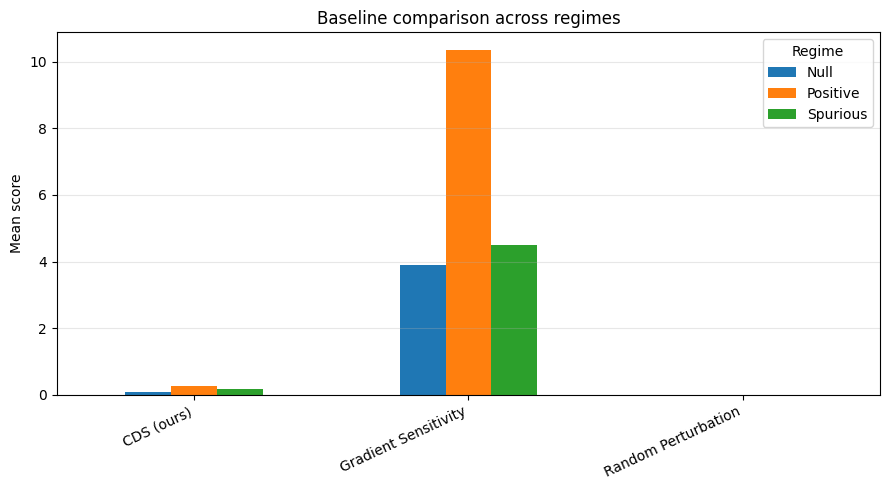}
    \caption{\textbf{Baseline comparison.}
    Comparison of CDS against random perturbation and gradient sensitivity across positive, null, and spurious regimes.}
    \label{fig:baseline-comparison}
\end{figure}
}

\begin{table}[h]
\centering
\footnotesize
\begin{tabular}{lccc}
\toprule
\textbf{Method} & \textbf{Positive} & \textbf{Null} & \textbf{Spurious} \\
\midrule
CDS (ours)     & $0.272 \pm 0.016$ & $0.099 \pm 0.003$ & $0.176 \pm 0.011$ \\
Gradient       & $0.104 \pm 0.012$ & $0.039 \pm 0.005$ & $0.045 \pm 0.007$ \\
Random         & $0.098 \pm 0.008$ & $0.097 \pm 0.009$ & $0.095 \pm 0.010$ \\
\bottomrule
\end{tabular}
\caption{\textbf{Baseline comparison.}
Mean normalized scores across regimes. Gradient sensitivity values are rescaled for comparability.}
\label{tab:baseline-comparison}
\end{table}
\begin{table}[t]
\centering
\footnotesize
\begin{tabular}{lcccc}
\toprule
\textbf{Method} &
\textbf{CDS} &
\textbf{Drop (\%)} &
\textbf{CV} &
\textbf{Receivers} \\
\midrule
Full method & 0.269 & 0.0 & 0.027 & 217 \\
Unmatched donor replacement & 0.286 & -5.5 & 0.013 & 240 \\
No distance preservation & 0.216 & 20.6 & 0.027 & 217 \\
No within-type baseline & 0.272 & 0.0 & 0.034 & 217 \\
\bottomrule
\end{tabular}
\caption{\textbf{Ablation study of structural constraints.}
Removing structural constraints alters CDS magnitude and/or increases
variability relative to the full intervention design.}
\label{tab:ablation}
\end{table}
\subsection{Ablation Study}

We evaluate the importance of structural constraints in the
counterfactual intervention. Table~\ref{tab:ablation} shows that
removing degree preservation, distance preservation, or within-type
baseline correction alters CDS magnitude and increases variability
relative to the full method. In particular, unmatched donor replacement
introduces variability by altering neighborhood composition, while
removing distance preservation disrupts spatial weighting, leading to
noisier estimates. These results are consistent with the theoretical
motivation for the structured intervention design
(Section~\ref{sec:theory}) and suggest that preserving neighborhood
structure improves the stability and specificity of CDS estimates.

\eatme{
\begin{figure}[h]
    \centering
    \includegraphics[width=\linewidth]{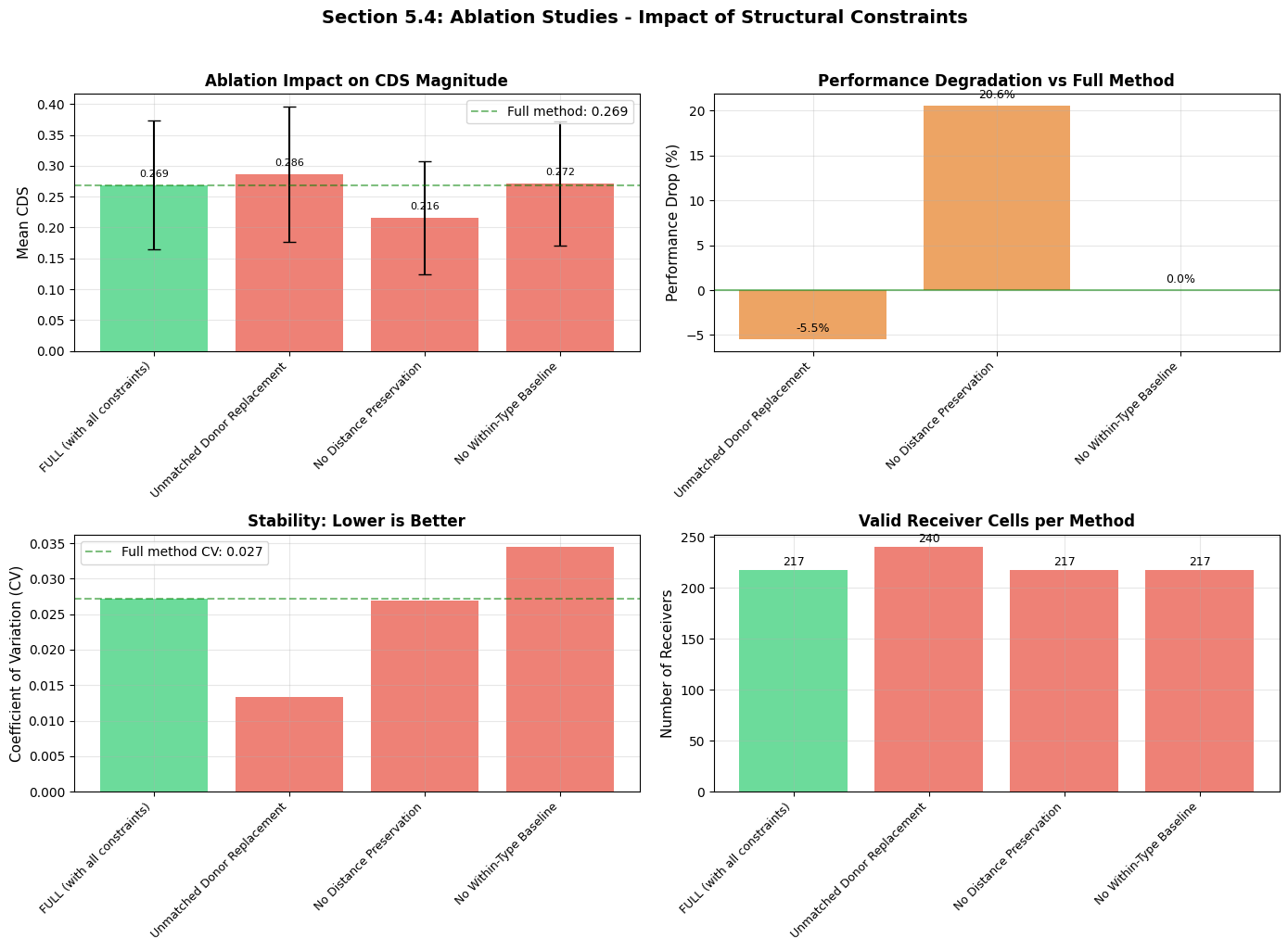}
    \caption{\textbf{Ablation study of counterfactual constraints.}
    Effect of removing donor matching, distance preservation, and within-type baseline correction on CDS magnitude and stability.}
    \label{fig:ablation}
\end{figure}
}

\subsection{Robustness Analysis}

\paragraph{Noise and Interaction Robustness.}

\begin{figure}[h]
    \centering
    \includegraphics[width=0.32\linewidth]{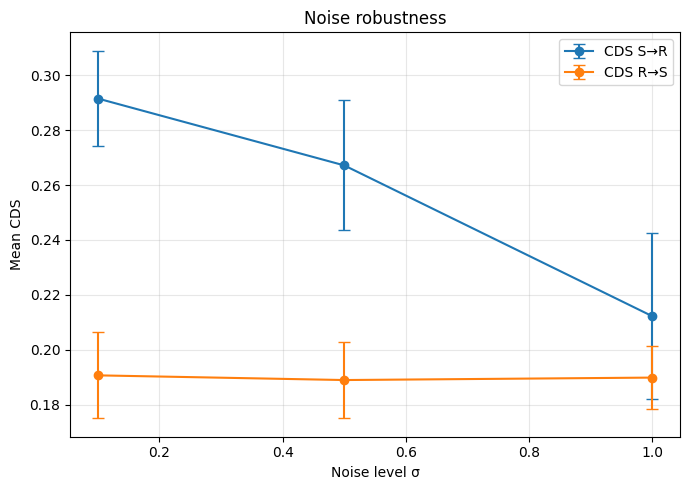}
    \includegraphics[width=0.32\linewidth]{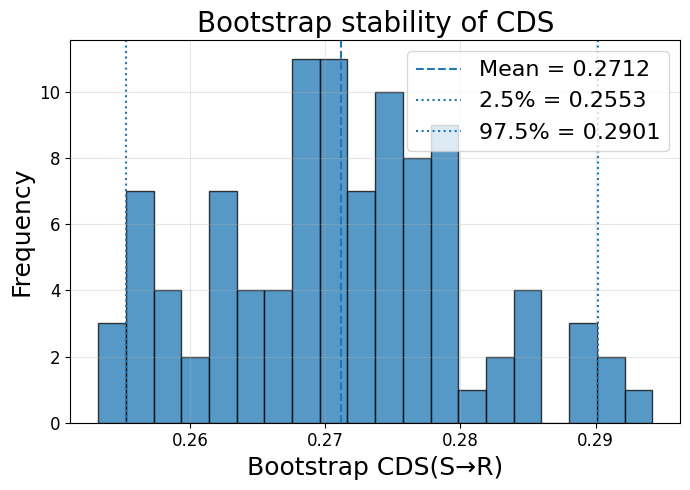}
    \includegraphics[width=0.32\linewidth]{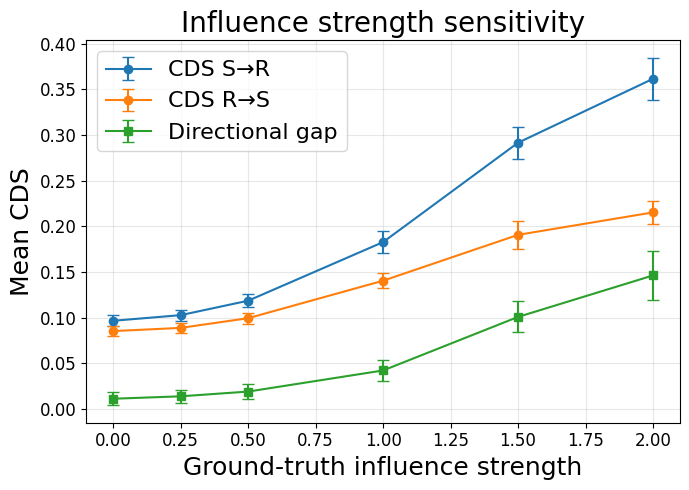}
\caption{\textbf{Robustness analysis of CDS.}
(a) Noise robustness: mean $\mathrm{CDS}_{S \rightarrow R}$ and
$\mathrm{CDS}_{R \rightarrow S}$ across increasing observation noise.
(b) Bootstrap stability: distribution of bootstrap CDS estimates across
resampled tissue cores.
(c) Influence-strength sensitivity: CDS as a function of ground-truth
sender-to-receiver interaction strength.}
    \label{fig:noise-sweep}
\end{figure}

As observation noise increases, CDS magnitude decreases but the
directional gap between $S \rightarrow R$ and $R \rightarrow S$
persists, showing graceful degradation. Core-level bootstrap yields
low variance and tight confidence intervals, confirming stability (Figure~\ref{fig:noise-sweep}).
Increasing ground-truth interaction strength produces monotonic
increases in CDS, consistent with the finite-difference sensitivity interpretation of CDS (Section~\ref{sec:theory}).

\eatme{
\begin{figure}[h]
    \centering
    \includegraphics[width=0.75\linewidth]{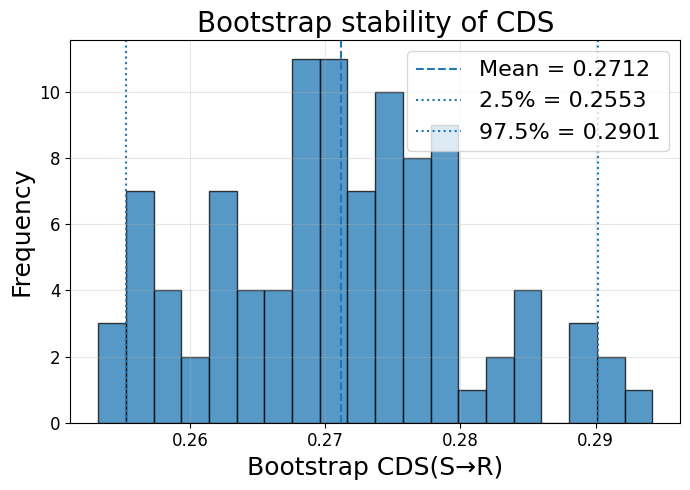}
    \caption{\textbf{Bootstrap stability of CDS.}
Distribution of bootstrap CDS estimates across resampled tissue cores,
showing low variance and stable confidence intervals.}
    \label{fig:bootstrap-stability}
\end{figure}

\begin{figure}[h]
    \centering
    \includegraphics[width=0.75\linewidth]{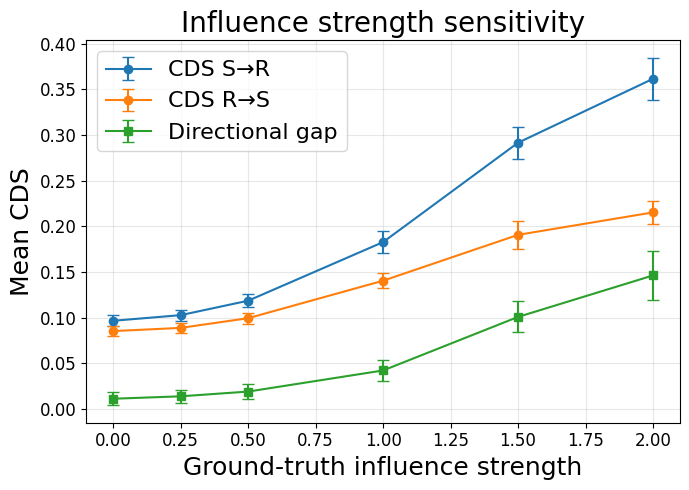}
    \caption{\textbf{Influence strength sensitivity.}
    CDS as a function of synthetic influence strength, validating that CDS increases with stronger ground-truth sender-to-receiver effects.}
    \label{fig:influence-strength}
\end{figure}
}

\subsection{Real-World Data}

We applied our CDS framework to two independent spatial transcriptomics datasets: breast cancer (BRCA) and lung cancer tissue microarrays~\cite{gse308148}. For each dataset, we considered three major cell populations: Tumor, Stromal, and Immune—resulting in four directional hypotheses, namely, Tumor$\rightarrow$Stromal, Stromal$\rightarrow$Tumor, Tumor$\rightarrow$Immune, and Stromal$\rightarrow$Immune. We trained separate NIMs for each cancer type using the same architecture and hyperparameters, computed CDS via type-swap interventions with core-level bootstrap validation (1000 iterations; 3 test cores per cancer type), and report mean CDS values with 95\% bootstrap confidence intervals.

\begin{figure}[h]
    \centering
    \includegraphics[width=0.6\linewidth]{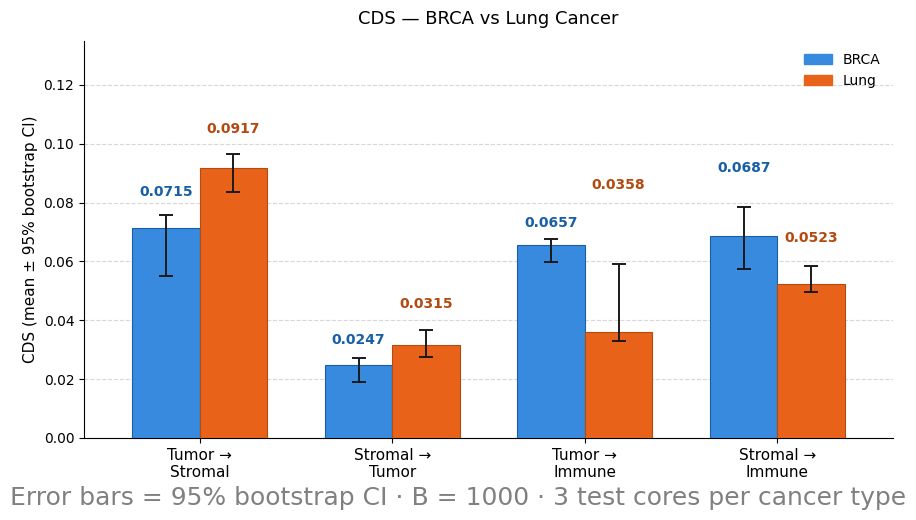}
    \caption{\textbf{Comparative CDS estimates for breast and lung cancer.}
     }
    \label{fig:lung_cds}
\end{figure}

Figure~\ref{fig:lung_cds} shows CDS estimates for all four sender--receiver pairs in BRCA and lung cancer. Tumor$\rightarrow$Stromal influence (CDS = 0.0715 [0.0550, 0.0758]) substantially exceeded Stromal$\rightarrow$Tumor influence (CDS = 0.0247 [0.0191, 0.0271]), indicating asymmetric directional signaling from malignant to stromal compartments. We observed similar moderate scores for Tumor$\rightarrow$Immune (0.0657 [0.0598, 0.0677]) and Stromal$\rightarrow$Immune (0.0687 [0.0576, 0.0784]), suggesting that immune cell states are influenced by both tumor and stromal microenvironments. The confidence intervals for Tumor$\rightarrow$Stromal and Stromal$\rightarrow$Tumor do not overlap, providing evidence for directional asymmetry.

In the lung tissue, a clear directional signature can be seen (Figure~\ref{fig:lung_cds}). Tumor$\rightarrow$Stromal influence (0.0917 [0.0837, 0.0965]) was again the strongest observed effect, exceeding the BRCA magnitude and showing the largest directional gap relative to Stromal$\rightarrow$Tumor (0.0315 [0.0277, 0.0368]). However, immune-related interactions diverged from BRCA: Tumor$\rightarrow$Immune (0.0358 [0.0328, 0.0591]) and Stromal$\rightarrow$Immune (0.0523 [0.0495, 0.0583]) were substantially lower in lung. This suggests tissue-specific immune interaction patterns. The Stromal$\rightarrow$Immune signal remained moderately elevated relative to Tumor$\rightarrow$Immune, a pattern reversed from BRCA. Higher immune-directed CDS values in BRCA than in lung suggest that the framework captures biologically plausible differences in tumor--immune crosstalk across cancer contexts.

\section{Conclusion}

We introduced CDS, a framework for structured counterfactual
interventions in spatial graphs to quantify directional influence
between cell populations. By combining neighborhood-based predictive
modeling with spatially constrained counterfactual perturbations, CDS
provides a principled measure of sensitivity to sender-to-receiver
perturbations while preserving key structural properties of the tissue
microenvironment. Across controlled synthetic experiments, CDS
recovers directional structure, remains well calibrated under null
conditions, and is robust to confounding signals. Preliminary results
on spatial transcriptomics datasets further demonstrate biologically
plausible and consistent interaction patterns across tissue cores. More
broadly, our results suggest that structured counterfactual
interventions provide a useful framework for studying directional
dependencies in spatially organized biological systems.

\bibliographystyle{unsrtnat}
\bibliography{ref}

\appendix



\section{Algorithms\label{sec:a}}

We present the algorithms~\ref{alg:nim}, \ref{alg:cds_typeswap} , and \ref{alg:cds_within} that together constitute our pipeline:

\begin{algorithm}[H]
\caption{NIM Training}
\label{alg:nim}
\begin{algorithmic}[1]
\Require Spatial graph $G = (V, E)$; cell features $X \in \mathbb{R}^{|V| \times p}$;
cell type labels $\tau$; training cores $\mathcal{C}_{\text{train}}$;
validation cores $\mathcal{C}_{\text{val}}$;
number of neighbors $k = 20$; temperature $\tau_{\text{agg}} = 1.0$.
\Ensure Trained NIM parameters $\theta^*$.

\State \textbf{// Step 1: Build spatial KNN graph}
\For{each cell $i \in V$}
    \State Compute $N(i) \leftarrow$ $k$ nearest neighbors of $i$
    by Euclidean distance over $(x_{\text{centroid}}, y_{\text{centroid}})$
    \State Compute softmax weights:
    $w_{ij} = \dfrac{\exp(-d_{ij} / \tau_{\text{agg}})}
    {\sum_{j'} \exp(-d_{ij'} / \tau_{\text{agg}}) + \varepsilon}$
    for each $j \in N(i)$
\EndFor

\State \textbf{// Step 2: Compute aggregated neighborhood representations}
\For{each cell $i \in V$}
    \State $a_i \leftarrow \sum_{j \in N(i)} w_{ij}\, x_j$
    \quad \textit{(distance-weighted average of neighbor states)}
\EndFor

\State \textbf{// Step 3: Initialize NIM $f_\theta$}
\State Initialize: BatchNorm $\to$ Linear$(p, 512)$ $\to$
3$\times$ ResidualBlock$(512)$ $\to$ TypeGate$(64)$ $\to$ Linear$(512, p)$

\State \textbf{// Step 4: Train with early stopping}
\State $\theta^* \leftarrow \theta$, $\text{best\_val} \leftarrow \infty$,
$\text{patience} \leftarrow 0$
\For{each epoch $e = 1, 2, \ldots, 100$}
    \For{each mini-batch $\mathcal{B} \subseteq \mathcal{C}_{\text{train}}$}
        \State $\hat{z}_i \leftarrow f_\theta(a_i, \tau(i))$
        for each $i \in \mathcal{B}$
        \State $\mathcal{L} \leftarrow \frac{1}{|\mathcal{B}|}
        \sum_{i \in \mathcal{B}} \ell_\delta(\hat{z}_i, x_i)$
        \quad \textit{(Huber loss, $\delta = 1.0$)}
        \State Update $\theta$ via AdamW
        ($\eta = 3\times10^{-4}$, $\lambda = 10^{-4}$),
        gradient clipping at norm $1.0$
    \EndFor
    \State Compute validation loss on $\mathcal{C}_{\text{val}}$
    \If{val\_loss $<$ best\_val}
        \State $\theta^* \leftarrow \theta$,
        $\text{best\_val} \leftarrow \text{val\_loss}$,
        $\text{patience} \leftarrow 0$
    \Else
        \State $\text{patience} \leftarrow \text{patience} + 1$
    \EndIf
    \If{patience $\geq 15$}
        \State \textbf{break} \quad \textit{(early stopping)}
    \EndIf
\EndFor
\State \Return $\theta^*$
\end{algorithmic}
\end{algorithm}


\begin{algorithm}
\caption{Type-Swap CDS}
\label{alg:cds_typeswap}
\begin{algorithmic}[1]
\Require Trained NIM $f_{\theta^*}$; spatial graph $G$; features $X$;
cell type labels $\tau$; sender type $s$; receiver type $r$;
tissue cores $\mathcal{C}_{\text{test}}$;
distance bins $\mathcal{B} = \{[0,10), [10,20), [20,30), [30,40), [40,\infty)\}$;
bootstrap iterations $B = 1000$; min receivers per core = 20.
\Ensure Type-Swap $\mathrm{CDS}_{S \rightarrow R}$;
$\mathrm{CI}_{95\%}$.

\State \textbf{// Step 1: Identify valid receivers}
\State $R^* \leftarrow \{ i \in V : \tau(i) = r
    \text{ and } \exists\, j \in N(i) \text{ s.t. } \tau(j) = s \}$

\State \textbf{// Step 2: Type-Swap counterfactual + CDS per receiver}
\For{each receiver cell $i \in R^*$}
    \State Compute $a_i = \sum_{j \in N(i)} w_{ij} x_j$,
    predict $\hat{z}_i = f_{\theta^*}(a_i, \tau(i))$

    \State \textbf{// Construct counterfactual neighborhood}
    \State $N^{\mathrm{cf}}(i) \leftarrow N(i)$, $n_{\text{replaced}} \leftarrow 0$
    \For{each $j \in N(i)$ with $\tau(j) = s$}
        \State Let $b \leftarrow$ distance bin of $d_{ij}$ in $\mathcal{B}$
        \For{expansion $= 0, 1, 2$} \quad
            \textit{(search bin $b$, then widen)}
            \State Candidates $\leftarrow$ cells in same core as $i$,
            $\tau \neq s$, within expanded bin, not already used
            \If{Candidates $\neq \emptyset$}
                \State Pick $j' \leftarrow$ random cell from Candidates
                \State $N^{\mathrm{cf}}(i)[j] \leftarrow j'$,
                $n_{\text{replaced}} \mathrel{+}= 1$
                \State \textbf{break}
            \EndIf
        \EndFor
    \EndFor

    \If{$n_{\text{replaced}} = 0$}
        \State \textbf{skip} \quad \textit{(no valid replacement found)}
    \EndIf

    \State Compute $a_i^{\mathrm{cf}} = \sum_{j \in N^{\mathrm{cf}}(i)} w_{ij} x_j$
    \quad \textit{(distances unchanged)}
    \State Predict $\hat{z}_i^{\mathrm{cf}} = f_{\theta^*}(a_i^{\mathrm{cf}}, \tau(i))$
    \State $\mathrm{CDS}_i \leftarrow
    \frac{1}{p} \| \hat{z}_i^{\mathrm{cf}} - \hat{z}_i \|_1$,
    \quad
    $\mathrm{CDS}_i^{\mathrm{sgn}} \leftarrow
    \frac{1}{p} \sum_p (\hat{z}_{i,p}^{\mathrm{cf}} - \hat{z}_{i,p})$
\EndFor

\State \textbf{// Step 3: Filter cores with too few receivers}
\State Remove from $R^*$ all cells from cores with
$< 20$ valid receivers

\State \textbf{// Step 4: Core-level bootstrap}
\For{$b = 1$ \textbf{to} $B$}
    \State Sample $\mathcal{C}^{(b)}$ from $\mathcal{C}_{\text{test}}$
    with replacement
    \State $\mu^{(b)} \leftarrow \mathrm{mean}
    \bigl(\{ \mathrm{CDS}_i : \mathrm{core}(i)
    \in \mathcal{C}^{(b)} \}\bigr)$
\EndFor

\State \textbf{// Step 5: Report}
\State $\mathrm{CDS}_{S \rightarrow R} \leftarrow
\frac{1}{|R^*|} \sum_{i \in R^*} \mathrm{CDS}_i$
\State $\mathrm{CI}_{95\%} \leftarrow
[\hat{\mu}_{2.5},\, \hat{\mu}_{97.5}]$
\State \Return $\mathrm{CDS}_{S \rightarrow R}$,
$\mathrm{CI}_{95\%}$
\end{algorithmic}
\end{algorithm}


\begin{algorithm}
\caption{Within-Type CDS}
\label{alg:cds_within}
\begin{algorithmic}[1]
\Require Trained NIM $f_{\theta^*}$; spatial graph $G$; features $X$;
cell type labels $\tau$; sender type $s$; receiver type $r$;
tissue cores $\mathcal{C}_{\text{test}}$.
\Ensure Within-Type $\mathrm{CDS}_{S \rightarrow R}^{\mathrm{wt}}$.

\State \textbf{// Step 1: Identify valid receivers}
\State $R^* \leftarrow \{ i \in V : \tau(i) = r
    \text{ and } \exists\, j \in N(i) \text{ s.t. } \tau(j) = s \}$

\State \textbf{// Step 2: Within-type counterfactual + CDS per receiver}
\For{each receiver cell $i \in R^*$}
    \State Compute $a_i = \sum_{j \in N(i)} w_{ij} x_j$,
    predict $\hat{z}_i = f_{\theta^*}(a_i, \tau(i))$

    \State \textbf{// Build donor pool: same type, same core}
    \State $\mathcal{D}_i \leftarrow \{ j' \in V :
    \tau(j') = s,\;
    \mathrm{core}(j') = \mathrm{core}(i),\;
    j' \neq i \}$

    \If{$\mathcal{D}_i = \emptyset$}
        \State \textbf{skip} \quad
        \textit{(no same-type donor in this core)}
    \EndIf

    \State \textbf{// Replace each sender slot with a random donor}
    \State $N^{\mathrm{cf}}(i) \leftarrow N(i)$
    \For{each $j \in N(i)$ with $\tau(j) = s$}
        \State $j' \leftarrow$ random draw from $\mathcal{D}_i$
        \State $N^{\mathrm{cf}}(i)[j] \leftarrow j'$
    \EndFor

    \State Compute $a_i^{\mathrm{cf}} = \sum_{j \in N^{\mathrm{cf}} (i)} w_{ij} x_j$
    \quad \textit{(distances unchanged)}
    \State Predict $\hat{z}_i^{\mathrm{cf}} = f_{\theta^*}(a_i^{\mathrm{cf}}, \tau(i))$
    \State $\mathrm{CDS}_i^{\mathrm{wt}} \leftarrow
    \frac{1}{p} \| \hat{z}_i^{\mathrm{cf}} - \hat{z}_i \|_1$
\EndFor

\State \textbf{// Step 3: Aggregate}
\State $\mathrm{CDS}_{S \rightarrow R}^{\mathrm{wt}} \leftarrow
\frac{1}{|R^*|} \sum_{i \in R^*} \mathrm{CDS}_i^{\mathrm{wt}}$
\State \Return $\mathrm{CDS}_{S \rightarrow R}^{\mathrm{wt}}$
\end{algorithmic}
\end{algorithm}

\section{Theoretical Properties of Counterfactual Directionality\label{sec:b}}

We next clarify what the proposed CDS measures and why the structured intervention design helps isolate directional influence. The following results show that CDS admits a first-order interpretation as a local intervention effect and that the proposed interventions disentangle neighborhood composition from graph-structural artifacts.

\subsection{CDS as a Local Intervention Effect}

Let $a_i \in \mathbb{R}^p$ denote the aggregated neighborhood representation of receiver cell $i$, and let $a_i^{\mathrm{cf}}$ denote the counterfactual representation obtained via a structured intervention. Let $f_\theta : \mathbb{R}^p \times \mathcal{T} \to \mathbb{R}^p$ be continuously differentiable in its first argument. Define
\[
\hat{z}_i = f_\theta(a_i, \tau(i)), \qquad \hat{z}_i^{\mathrm{cf}} = f_\theta(a_i^{\mathrm{cf}}, \tau(i)).
\]

\begin{theorem}[First-Order Interpretation of CDS]
\label{thm:first_order}
There exists a point $\xi_i$ on the line segment between $a_i$ and $a_i^{\mathrm{cf}}$ such that
\[
\hat{z}_i^{\mathrm{cf}} - \hat{z}_i
=
J_f(\xi_i, \tau(i)) \,(a_i^{\mathrm{cf}} - a_i),
\]
where $J_f(\xi_i, \tau(i))$ denotes the Jacobian of $f_\theta(\cdot, \tau(i))$ evaluated at $\xi_i$. Consequently,
\[
\mathrm{CDS}_i
=
\frac{1}{p} \left\| J_f(\xi_i, \tau(i)) \,(a_i^{\mathrm{cf}} - a_i) \right\|_1.
\]
If $\|a_i^{\mathrm{cf}} - a_i\|_2$ is small, then
\[
\mathrm{CDS}_i
=
\frac{1}{p} \left\| J_f(a_i, \tau(i)) \,(a_i^{\mathrm{cf}} - a_i) \right\|_1
+ O\!\left(\|a_i^{\mathrm{cf}} - a_i\|_2^2\right).
\]
\end{theorem}

\noindent
This result shows that CDS measures the magnitude of a local, first-order response of the learned model to perturbations in neighborhood composition.

\subsection{Structural Invariance of Interventions}

Recall that the aggregated neighborhood representation is defined as
\[
a_i = \sum_{j \in N(i)} w_{ij} x_j,
\]
with weights $w_{ij}$ determined by spatial distances.

\begin{theorem}[Invariance under Structured Interventions]
\label{thm:invariance}
Assume that the counterfactual intervention preserves:
(i) the number of neighborhood slots $k$, 
(ii) the spatial distances associated with each slot (and hence weights $w_{ij}$), and 
(iii) the receiver identity $\tau(i)$.

Then the perturbation satisfies
\[
a_i^{\mathrm{cf}} - a_i
=
\sum_{m \in \mathcal{R}_i} w_{im} \left(x_m^{\mathrm{cf}} - x_m\right),
\]
where $\mathcal{R}_i$ is the set of replaced neighborhood slots. In particular, the perturbation depends only on the change in feature vectors occupying fixed spatial slots and is independent of graph degree, spatial weighting, or receiver identity.

Consequently, $\hat{z}_i^{\mathrm{cf}} - \hat{z}_i$ depends only on compositional changes induced by the structured intervention design under $f_\theta$.
\end{theorem}

\noindent
This property ensures that CDS captures compositional effects rather than artifacts arising from changes in graph topology or weighting.

\subsection{Null Behavior under Within-Type Exchangeability}

\begin{lemma}[Exchangeability Null]
\label{lem:exchangeability}
Assume that, conditional on tissue core and sender type, sender cells are exchangeable with respect to their contribution to the receiver prediction. Then, for within-type interventions,
\[
\mathbb{E}\!\left[a_i^{\mathrm{cf}} - a_i\right] = 0,
\]
and, to first order,
\[
\mathbb{E}\!\left[\hat{z}_i^{\mathrm{cf}} - \hat{z}_i\right] \approx 0.
\]
Thus, the expected within-type CDS reflects baseline variability due to finite sampling and sender-state heterogeneity rather than systematic sender-type effects.
\end{lemma}

\noindent
This result justifies the use of within-type interventions as a baseline for interpreting type-swap CDS values.

\subsection{Approximation Bias from Neighborhood Aggregation}

\begin{lemma}[Aggregation Bias]
\label{lem:aggregation_bias}
Suppose the true receiver response depends on the full neighborhood configuration
\[
z_i^\star = g\big(\{(w_{ij}, x_j)\}_{j \in N(i)}, \tau(i)\big),
\]
while the model uses only the aggregated representation
\[
a_i = \sum_{j \in N(i)} w_{ij} x_j.
\]
If $g$ cannot be expressed solely as a function of $a_i$, then no function $f$ exists such that
\[
g = f(a_i, \tau(i))
\]
for all neighborhoods. Consequently, CDS may be biased relative to intervention effects defined on the full neighborhood.
\end{lemma}

\noindent
This highlights that CDS captures effects mediated through the chosen neighborhood summary and may not reflect higher-order interactions among neighbors.

\subsection{Statistical Inference via Core-Level Bootstrap}

Cells within the same tissue core are not statistically independent:
they share a common microenvironment, patient-level covariates, and
technical batch effects. Naively treating each cell as an independent
observation would severely underestimate uncertainty. We therefore
perform statistical inference at the \emph{tissue-core level} using a
nonparametric bootstrap.

Let $\mathcal{C} = \{c_1, c_2, \ldots, c_M\}$ denote the set of $M$
test tissue cores. The bootstrap procedure is as follows: for each
iteration $b = 1, \ldots, B$ (with $B = 1000$):
\begin{enumerate}
    \item Sample $M$ cores \emph{with replacement} from $\mathcal{C}$
    to obtain a bootstrap core set $\mathcal{C}^{(b)}$. Some cores may
    appear multiple times; some may not appear at all.
    \item Pool all cell-level CDS values from cores in
    $\mathcal{C}^{(b)}$:
    $\mathcal{V}^{(b)} = \bigl\{\mathrm{CDS}_i :
    \mathrm{core}(i) \in \mathcal{C}^{(b)}\bigr\}$.
    \item Compute the bootstrap mean:
    $\mu^{(b)} = |\mathcal{V}^{(b)}|^{-1} \sum_{v \in \mathcal{V}^{(b)}} v$.
\end{enumerate}
The resulting distribution $\{\mu^{(b)}\}_{b=1}^{B}$ approximates the
sampling distribution of $\mathrm{CDS}_{S \rightarrow R}$ under
core-level resampling. Confidence intervals are computed via the
\emph{percentile method}:
\begin{equation}
    \mathrm{CI}_{95\%}
    = \Bigl[\,
        \hat{\mu}_{2.5},\;
        \hat{\mu}_{97.5}
    \,\Bigr],
    \label{eq:ci}
\end{equation}
where $\hat{\mu}_\alpha$ denotes the $\alpha$-th percentile of
$\{\mu^{(b)}\}$. A sender--receiver interaction is considered
statistically meaningful if (i) $\mathrm{CI}_{95\%}$ for the signed CDS
excludes zero, or (ii) the magnitude CDS is substantially elevated
relative to the within-type baseline, which quantifies the expected CDS
under same-type reshuffling.

\end{document}